\documentclass{article}

\usepackage{microtype}
\usepackage{graphicx}
\usepackage[caption=false]{subfig}
\usepackage{booktabs} 
\usepackage{enumitem}
\usepackage{hyperref}
\usepackage{amsfonts}
\usepackage{amsmath}
\usepackage{bbm}
\usepackage{scalerel}
\usepackage{dblfloatfix}
\usepackage{amssymb}
\usepackage{pifont}
\newcommand{\cmark}{\ding{51}}%
\newcommand{\xmark}{\ding{55}}%
\usepackage{wrapfig}
\usepackage[dvipsnames]{xcolor}

\usepackage{array}
\newcolumntype{L}{>{\centering\arraybackslash}m{0.2\columnwidth}}
\usepackage{soul}

\usepackage[accepted]{icml2021}
\renewcommand{\algorithmiccomment}[1]{\texttt{{\color{blue} #1}}}
\icmltitlerunning{Cross-domain Imitation from Observations}

\begin{document}

\twocolumn[
\icmltitle{Cross-domain Imitation from Observations}



\icmlsetsymbol{equal}{*}
\icmlsetsymbol{work}{$\dagger$}

\begin{icmlauthorlist}
\icmlauthor{Dripta S. Raychaudhuri}{equal,ucr}
\icmlauthor{Sujoy Paul}{equal,goo,work}
\icmlauthor{Jeroen van Baar}{merl}
\icmlauthor{Amit K. Roy-Chowdhury}{ucr}
\end{icmlauthorlist}

\icmlaffiliation{ucr}{University of California, Riverside }
\icmlaffiliation{goo}{Google Research }
\icmlaffiliation{merl}{Mitsubishi Electric Research Laboratories }

\icmlcorrespondingauthor{Dripta S. Raychaudhuri}{draychaudhuri@ece.ucr.edu}

\icmlkeywords{Imitation learning, Domain adaptation}

\vskip 0.3in
]



\printAffiliationsAndNotice{\icmlEqualContribution \work} 

\begin{abstract}
Imitation learning seeks to circumvent the difficulty in designing proper reward functions for training agents by utilizing expert behavior. With environments modeled as Markov Decision Processes (MDP), most of the existing imitation algorithms are contingent on the availability of expert demonstrations in the \textit{same} MDP as the one in which a new imitation policy is to be learned. In this paper, we study the problem of how to imitate tasks when there exists discrepancies between the expert and agent MDP. These discrepancies across domains could include differing dynamics, viewpoint or morphology; we present a novel framework to learn correspondences across such domains. Importantly, in contrast to prior works, we use \textit{unpaired} and \textit{unaligned} trajectories containing \textit{only states} in the expert domain, to learn this correspondence. We utilize a cycle-consistency constraint on both the state space and a domain agnostic latent space to do this. In addition, we enforce consistency on the temporal position of states via a normalized position estimator function, to align the trajectories across the two domains. Once this correspondence is found, we can directly transfer the demonstrations on one domain to the other and use it for imitation. Experiments across a wide-variety of challenging domains demonstrate the efficacy of our approach.
\end{abstract}


\section{Introduction}
\begin{figure}
    \centering
    \includegraphics[scale=0.35]{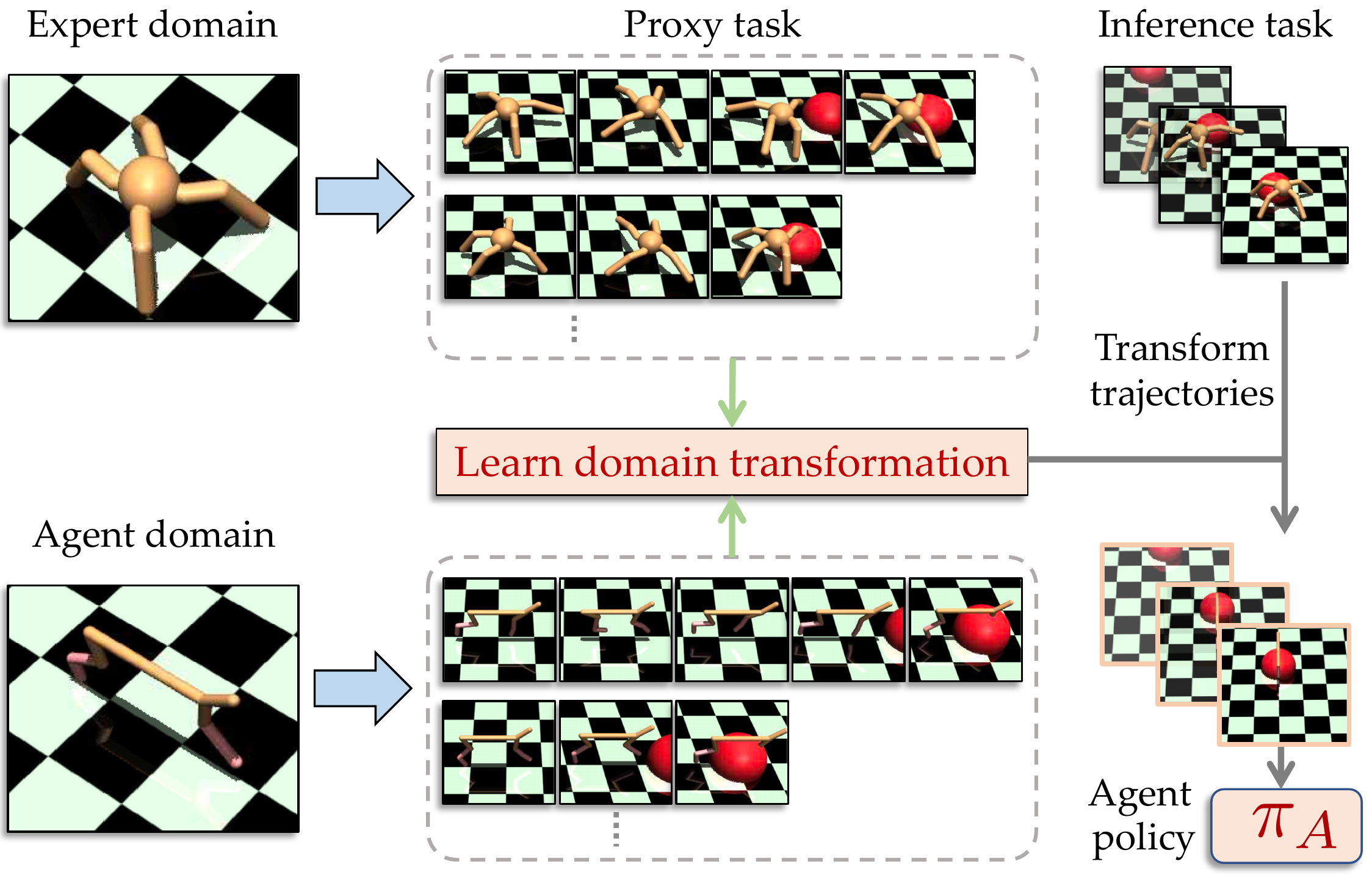}
    \caption{\textbf{Problem overview.} Cross-domain Imitation from Observation (\texttt{xDIO}) entails learning from experts which are different from the agent. Here, the expert is a 4-legged Ant, while the agent is a HalfCheetah. We learn a domain transformation function from unpaired, unaligned, state-only trajectories from a set of proxy tasks and utilize it to imitate the expert on the given inference task.}
    \label{fig:overview}
    \vskip -0.2in
\end{figure}

Humans possess the innate ability to quickly pick up a new behavior by simply observing others performing the same skill. Not only are we able to learn from demonstrations coming from a third-person point of view, we are also capable of imitation from experts who are morphologically different or have different embodiments - as evidenced by a child imitating an adult with different bio-mechanics \cite{jones2009development}. Previous works in neuroscience \cite{rizzolatti2004mirror,marshall2015body} have attributed this to the human capacity of learning structure preserving domain correspondences via an invariant feature space \cite{umilta2008pliers}, which allows us to reconstruct the observed behavior in the self-domain. While imitation learning algorithms \cite{ho2016generative,ross2011reduction} are successful, to some extent, in endowing autonomous agents with this ability to imitate expert behavior, they impose the somewhat unrealistic requirement that the demonstrations must come from the same domain, whether that be first-person viewpoint, same morphology or similar dynamics. The question then arises: \textit{can we perform imitation learning which can overcome all such domain discrepancies?}

Prior work on bridging domain disparities in imitation learning have focused on each of these differences in isolation: morphology \cite{gupta2017invariant}, dynamics \cite{gangwani2019state} and viewpoint mismatch \cite{stadie2017tpil,sharma2019third,liu2018imitation}. These works \cite{gupta2017invariant,liu2018imitation,sharma2019third} utilize \textit{paired, time-aligned} demonstrations from both domains, on a set of \textit{proxy} tasks, to first build a correspondence map across the domains and then perform an extra reinforcement learning (RL) step for learning the final policy on the given task. This limits their applicability since paired demonstrations are rarely available and RL procedures are expensive. 

Recently, \cite{kim2020domain} proposed a general framework which can perform imitation across a wide array of such discrepancies from \textit{unpaired, unaligned} demonstrations. However, they require \textit{expert actions}, such as the exact kinematic forces, in order to learn a domain correspondence and assume availability of an expert policy which is utilized in an \textit{interactive learning} setting. This is distinctly different to how humans imitate: we are capable of learning behaviors solely from observations/states, without access to underlying actions. Furthermore, continuously querying the expert might be onerous in several situations. Thus, we require a mechanism for learning policies from observation alone, where the expert demonstrations can originate in a domain which is different from the agent domain and access to the expert is limited. We define this setting as \textit{Cross Domain Imitation from Observation} (\texttt{xDIO}). 



In this work, we propose a novel framework to tackle the \texttt{xDIO} problem, encompassing morphological, viewpoint and dynamics mismatch. We follow a two-step approach (see Fig. \ref{fig:overview}), where we first learn a transformation across the domains using the proxy tasks \cite{gupta2017invariant}, followed by a transfer process and subsequent learning of the policy. Importantly, in contrast to previous work, we use \textit{unpaired} and \textit{unaligned} trajectories containing \textit{only states} on the expert domain trajectories, to learn this transformation. Additionally, we do not assume any access to the expert policy or the expert domain except for the given demonstrations. To learn the state correspondences, we jointly minimize a divergence between the transition distributions in the state space as well as in the latent space between the expert and the agent proxy task trajectories, while learning to translate between the two domains with the unpaired data via cycle-consistency \cite{zhu2017unpaired}. However, solely learning with such state cycle-consistency may only result in local alignment, and lead to difficulties in optimizing for complex environments. Thus, to impose global alignment, we enforce additional consistency on the temporal position of states across the two domains. This ensures that when a state is mapped from one domain to the other, the degree of completion associated by being in that state remains unchanged. Having learnt this mapping on the proxy tasks, we transfer demonstrations for a new \textit{inference} task from the expert to the agent domain, which are subsequently utilized to learn a policy via imitation.

Experiments on a wide array of domains that encompass dynamics, morphological and viewpoint mismatch, demonstrate the feasibility of learning domain correspondences from unpaired and unaligned state-only demonstrations. The primary contributions of this work are as follows:

\vspace{0.5mm}
(a) We propose an algorithm for cross-domain imitation learning by learning transformations across domains, modeled as Markov Decision Processes (MDP), from \textit{unpaired, unaligned, state-only} demonstrations, thereby ameliorating the need for costly paired, aligned data.\\
(b) Unlike previous work, neither do we utilize any costly RL procedure, nor do we require interactive querying of an expert policy. \\
(c) We adopt multiple tasks in the MuJoCo physics engine~\cite{todorov2012mujoco}, and show that our framework can find correspondences and align two domains across \textit{different viewpoints, dynamics and morphologies}.


\section{Related Works}
\textbf{Imitation learning.} Imitation learning \cite{schaal1999imitation} uses a set of expert demonstrations to learn a policy which successfully mimics the expert. A common approach is behavioral cloning (BC) \cite{pomerleau1989alvinn,bojarski2016end}, which amounts to learning to mimic the expert demonstrations via supervised learning. 
Inverse reinforcement learning (IRL) is another approach, where one seeks to learn a reward function that explains the demonstrated actions \cite{ho2016generative,abbeel2004apprenticeship,ziebart2008maximum}. Recent works \cite{torabi2018behavioral,yang2019imitation,paul2019learning} extend imitation learning to state-only demonstrations, where expert actions are not observed - this opens up the possibility of using imitation in robotics and learning from weak-supervision sources such as videos. Unlike these approaches, our work tackles the problem of imitation from state-only demonstrations coming from a \textit{different} domain.


\textbf{Domain transfer in reinforcement learning.} Transfer in the reinforcement learning setting has been attempted by a wide array of works \cite{taylor2009transfer}. \cite{ammar2011reinforcement} manually define a common state space between MDPs and use it to learn a mapping between states. Unsupervised manifold alignment is used in \cite{ammar2015unsupervised} to learn a linear map between states with similar local geometric properties. However, they assume the
existence of hand-crafted features along with a distance metric between them, which limits its applicability. Recent works in transfer learning across mismatches in embodiment \cite{gupta2017invariant} and viewpoint \cite{liu2018imitation,sharma2019third}, obtain state correspondences from an proxy task set comprising paired, time-aligned demonstrations and use them to learn a state map or a state encoder to a domain invariant feature space. \cite{kim2020domain} proposed a framework which can learn a map across domains from unpaired, unaligned demonstrations. However, they require expert actions to train the framework, along with access to an online expert. Furthermore, most of these approaches \cite{gupta2017invariant, liu2018imitation} utilize an RL step which incurs additional computational cost. In contrast to these methods, our approach learns an MDP structure preserving state map from unpaired, unaligned demonstrations without requiring access to expert actions, additional RL or online experts. Viewpoint agnostic imitation has also been tackled in \cite{stadie2017tpil}, where a combination of adversarial learning \cite{ho2016generative} and domain confusion  \cite{tzeng2014deep} is used to learn a policy without an proxy set. However, it fails to account for large variations in viewpoint, in addition to sub-optimal trajectories from the expert domain. From a theoretical perspective, our approach aligns with the objective of MDP homomorphisms \cite{ravindran2004algebraic}. Similar ideas are explored in learning MDP similarity metric via bisimulation \cite{ferns2011bisimulation} and Boltzmann machine reconstruction error \cite{ammar2014automated}. However, these works find homomorphisms within an MDP and do not provide ways to discover homomorphisms across MDPs.

\textbf{Cycle-consistency.} Our work draws inspiration from the literature on cycle-consistency \cite{zhu2017unpaired,hoffman2018cycada,smith2019avid}. CycleGAN \cite{zhu2017unpaired} introduced cycle-consistency to learn bidirectional transformations between domains via
Generative Adversarial Networks \cite{goodfellow2014generative} for unpaired image-to-image translation. This was extended to domain adaptation in \cite{hoffman2018cycada}. Similar techniques are applied in sim-to-real transfer \cite{ho2020retinagan,gamrian2019transfer}. Recently, \cite{rao2020rl} propose RL-CycleGAN to perform sim-to-real transfer by adding extra supervision from the Q-value function. Unlike these works, which are restricted on visual alignments, we propose to learn alignments across differing dynamics/morphology.

\begin{table}
\centering
\caption{Comparison to prior work using attributes demonstrated in the paper. \texttt{xDIO} satisfies all the criteria desired in a holistic domain adaptive imitation framework.}
\vskip 0.05in
\resizebox{0.47\textwidth}{!}{
\begin{tabular}{@{}cLLLL@{}}
\toprule
\textsc{Method} & \textsc{Unpaired trajectories} & \textsc{Only states} & \textsc{No online expert} & \textsc{No RL} \\ \midrule
IF \cite{gupta2017invariant}
& \color{red}\xmark
& \color{ForestGreen}\cmark
& \color{ForestGreen}\cmark
& \color{red}\xmark \\
DAIL \cite{kim2020domain}
& \color{ForestGreen}\cmark
& \color{red}\xmark
& \color{red}\xmark
& \color{ForestGreen}\cmark \\
Ours  
& \color{ForestGreen}\cmark
& \color{ForestGreen}\cmark
& \color{ForestGreen}\cmark
& \color{ForestGreen}\cmark \\
\bottomrule
\end{tabular}
}
\vskip -0.2in
\end{table}
\section{Problem Setting} \label{sec:setting}
Before formally defining the \texttt{xDIO} problem, we first lay the groundwork in terms of notation. Following \cite{kim2020domain}, we define a domain as a tuple $\left(\mathcal{S}, \mathcal{A}, \mathcal{P}, \mathcal{P}_0\right)$, where $\mathcal{S}$ denotes the state space, $\mathcal{A}$ is the action space, $\mathcal{P}$ is the dynamics or transition function, and $\mathcal{P}_0$ is the initial distribution over the states. Given an action $a\in\mathcal{A}$, the next state is governed by the transition dynamics as $s'\sim\mathcal{P}(s'|s,a)$. An infinite horizon Markov Decision Process (MDP) is defined subsequently by including a reward function $r:\mathcal{S}\times\mathcal{A}\rightarrow\mathbb{R}$, and a discount factor $\gamma\in[0, 1]$ to the domain tuple. Thus, while the domain typifies only the agent morphology and the dynamics, augmenting the domain with a reward and discount factor describes an MDP for a particular task. We define an MDP in some domain $x$ for a task $\mathcal{T}$ as $\mathcal{M}_x^\mathcal{T}=\left(\mathcal{S}_x,\mathcal{A}_x,\mathcal{P}_x,r_x^\mathcal{T},\gamma_x^\mathcal{T},\mathcal{P}_{0x}\right)$. A policy is a map $\pi_x^\mathcal{T}:\mathcal{S}_x\rightarrow\mathcal{B}(\mathcal{A}_x)$, where $\mathcal{B}$ is the set of all probability measures on $A_x$.
A trajectory corresponding to the task $\mathcal{T}$ in domain $x$ is a sequence of states $\eta_{\mathcal{M}_x^\mathcal{T}}=\{s_x^0,s_x^1,\dots,s_x^{H_{\eta}}\}$, where $H_\eta$ denotes the length of the trajectory. We denote $\mathcal{D}_{\mathcal{M}_x^\mathcal{T}}=\{\eta^i_{\mathcal{M}_x^\mathcal{T}}\}_{i=1}^N$ to be a set of such trajectories. In our work, we consider two domains: expert and agent, indicated by $\mathcal{M}_E^\mathcal{T}$ and $\mathcal{M}_A^\mathcal{T}$ respectively.


The objective of \texttt{xDIO} is to learn an optimal policy $\pi_A^\mathcal{T}$ in the agent domain, given state-only demonstrations $\mathcal{D}_{\mathcal{M}_E^\mathcal{T}}$ in the expert domain. In this paper, we propose to first learn a transformation $\psi:\mathcal{S}_E\rightarrow\mathcal{S}_A$ between the domains and then leverage $\psi$ to imitate from the expert demonstrations. Following prior work \cite{gupta2017invariant,liu2018imitation,kim2020domain}, we assume access to a dataset consisting of expert-agent trajectories for $M$ different \emph{proxy} tasks:  $\mathcal{D}=\{(\mathcal{D}_{\mathcal{M}_E^{\mathcal{T}_j}},\mathcal{D}_{\mathcal{M}_A^{\mathcal{T}_j}})\}_{j=1}^M$. 
Proxy tasks encompass simple primitive skills in both domains and are different from the inference task $\mathcal{T}$, for which we wish to learn the policy. 

We relax certain assumptions made in previous work, which are critical for real-world applications. Firstly, the trajectories derived from proxy tasks are not paired, i.e., time-aligned trajectories do not exist in $\mathcal{D}$. This is crucial in real-world cases, as the tasks may not be executed at the same rate in different domains. Secondly, expert actions are not observed: such actions are difficult to obtain in various scenarios such as videos of humans performing some task. Finally, we train in an offline fashion and do not require any expert policy for interactive querying, to guide the learning process, beyond the provided demonstrations.





Once the domain transformation function $\psi$ is learnt, we use it to translate the expert domain trajectories  $\mathcal{D}_{\mathcal{M}_E^\mathcal{T}}$, for the inference task $\mathcal{T}$, to the agent domain to obtain $\hat{\mathcal{D}}_{\mathcal{M}_A^\mathcal{T}}$. An inverse dynamics model $\mathcal{I}_A:\mathcal{S}_A\times\mathcal{S}_A\rightarrow\mathcal{A}_A$ is then learnt to augment these translated trajectories with actions, similar to \cite{torabi2018behavioral}. These are subsequently used to learn the policy $\pi_A^\mathcal{T}$ via imitation learning. 

\section{Method}
\label{sec:method}
A crucial characteristic of a good domain transformation $\psi$ lies in MDP dynamics preservation. In our framework, we enforce this from both the local and global perspectives. For local alignment, we aim to ensure that optimal state transitions in $\mathcal{M}_E^\mathcal{T}$ map to optimal transitions in $\mathcal{M}_A^\mathcal{T}$. Our proposed method achieves this local alignment by matching the state-transition distributions defined for the true and transferred trajectories on the proxy tasks in an adversarial manner, while maintaining cycle-consistency. A latent space is learned via a mutual information objective to only preserve task-specific information. On the other hand, a learned temporal position function aims to enforce consistency on the temporal position of the states across the two domains to ensure global alignment. In the following parts, we define each of these components in more detail.

\subsection{Local alignment via distribution matching}

\noindent \textbf{State cycle-consistency.} 
We seek to map optimal transitions in the expert domain to the agent domain, and propose to learn domain transformation $\psi$ such that the state transition distribution is matched over the trajectories derived from the proxy tasks. We utilize adversarial training to accomplish this. Given unpaired samples $\{(s_E^{t}, s_E^{t+1})\} \in \mathcal{D}_{\mathcal{M}_E^{\mathcal{T}_j}}$ and $\{(s_A^{t}, s_A^{t+1})\} \in \mathcal{D}_{\mathcal{M}_A^{\mathcal{T}_j}}$ drawn from the $j^{th}$ proxy task, the function $\psi$ is learned in an adversarial manner with a discriminator $D_A^j$, where $\psi$ tries to map $(s_E^{t}, s_E^{t+1})$ onto the distribution of $(s_A^{t}, s_A^{t+1})$, while $D_A^j$ tries to distinguish translated samples $\left(\psi(s_E^{t}),\psi(s_E^{t+1})\right)$ against real samples $(s_A^{t}, s_A^{t+1})$:
\begin{align} \label{eq:cyc_gan}
    \min_{\psi} & \max_{D_A^j} \mathcal{L}_{adv}^j =  \mathbb{E}_{\scaleto{(s_A^{t}, s_A^{t+1})\sim \mathcal{D}_{\mathcal{M}_A^{\mathcal{T}_j}}}{12pt}} \bigg[\log D_A^j(s_A^{t}, s_A^{t+1})\bigg] \nonumber \\ 
    + &\mathbb{E}_{\scaleto{(s_E^{t}, s_E^{t+1})\sim \mathcal{D}_{\mathcal{M}_E^{\mathcal{T}_j}}}{12pt}} \bigg[\log (1-D_A^j(\psi(s_E^{t}), \psi(s_E^{t+1})))\bigg]
\end{align}


Solely optimizing this adversarial loss can lead to the model mapping the same set of states to any random permutation of states in the agent domain, where any of the learned mappings can induce an output distribution that matches the agent state transition distribution. Following \cite{zhu2017unpaired}, we introduce cycle consistency as a means to control this undesired effect. We learn another state map in the opposite direction $\phi:\mathcal{S}_A \rightarrow \mathcal{S}_E$ by optimizing an adversarial loss, $\min_{\phi}\max_{D_E^j} \mathcal{L}^j_{adv}$, with a discriminator $D_E^j$. Cycle consistency is then imposed as:
\begin{multline}\label{eq:cyc_con}
    \min_{\psi,\phi} \mathcal{L}_{cyc}^j = \mathbb{E}_{\scaleto{s_E\sim \mathcal{D}_{\mathcal{M}_E^{\mathcal{T}_j}}}{10pt}}\left[\Vert\phi\circ\psi(s_E)-s_E\Vert_2^2\right] + \\
    \mathbb{E}_{\scaleto{s_A\sim \mathcal{D}_{\mathcal{M}_A^{\mathcal{T}_j}}}{10pt}}\left[\Vert\psi\circ\phi(s_A)-s_A\Vert_2^2\right]
\end{multline}

\noindent \textbf{Domain invariant latent space.}
To incentivize $\psi,\phi$ to generalize beyond proxy tasks, we use an encoder-decoder structure for the transformation function $\psi$. Concretely, $\psi=\mathtt{D}_{E}\circ\mathtt{E}_{E}$, where $\mathtt{E}_{E}:\mathcal{S}_E\rightarrow\mathcal{Z}$ represents an encoder which maps a state in the expert domain to a domain agnostic latent space $\mathcal{Z}$, while $\mathtt{D}_{E}:\mathcal{Z}\rightarrow\mathcal{S}_A$ represents the decoding function.
$\phi=\mathtt{D}_{A}\circ\mathtt{E}_{A}$ is defined similarly via the same latent space $\mathcal{Z}$. Prior work \cite{gupta2017invariant} has explored learning such invariant spaces, but use paired data from both domains, which is a very strong and often unrealistic assumption, as explained above. Inspired from work based on information theoretic objectives \cite{eysenbach2018diversity,wan2020mutual}, we learn the latent space by minimizing the mutual information between the domain and the latent transitions:
\begin{equation}\label{eq:MI}
    \min_{\mathtt{E}_{E},\mathtt{E}_{A}} I\left(d;(z^t,z^{t+1})\right)
\end{equation}
where $(z^t,z^{t+1})$ denotes an encoded transition from either of the domains. Minimizing the mutual information between the domain ($\Delta=\{E,A\}$) and the encoded latent transition for the \textit{same} proxy task will result in a latent space which encodes the \textit{task-specific} information and filters out the \textit{domain-specific} nuances.

Note that we can decompose the mutual information term as $I\left(\Delta;(z^t,z^{t+1})\right)=H(\Delta)-H(\Delta|(z^t,z^{t+1}))$, where $H(\cdot)$ denotes the entropy. Thus, our objective in Equation \ref{eq:MI} reduces to just maximizing the conditional entropy $H(\Delta|(z^t,z^{t+1}))$. Due to intractability of this expression \cite{alemi2016deep,poole2019variational}, we optimize a variational lower bound $\mathbb{E}_{d\sim\Delta,(s_d^{t}, s_d^{t+1})\sim \mathcal{D}_{\mathcal{M}_d^{\mathcal{T}_j}}} \left[-\log q^j\left(d|(z^t,z^{t+1})\right)\right]$ instead, where $q^j$ denotes a variational distribution which approximates the true posterior.


Here, $q^j$ is parameterized as a discriminator which outputs the probability that the generated transition comes from domain $d$ for the $j$th proxy task. Maximizing this objective over the encoder parameters ensures that the discriminator is maximally confused and the latent transitions for the task, coming from both domains, are well aligned. The overall objective is as follows:
\begin{multline}
    \min_{q^j} \max_{\mathtt{E}_E,\mathtt{E}_A} \mathcal{L}_{MI} = \mathbb{E}_{d\sim\Delta,(s_d^{t}, s_d^{t+1})\sim \mathcal{D}_{\mathcal{M}_d^{\mathcal{T}_j}}} \left[-\log q^j\left(d|\right.\right.\\ \left.\left.(z^t,z^{t+1})\right)\right]
\end{multline}
Additionally, we enforce 
consistency in the latent embedding to further constrain the learnt mapping:
\begin{multline}\label{eq:emb_con}
    \min_{\psi,\phi} \mathcal{L}_{z}^j = \mathbb{E}_{\scaleto{s_E\sim \mathcal{D}_{\mathcal{M}_E^{\mathcal{T}_j}}}{10pt}}\left[\Vert\mathtt{E}_A\circ\psi(s_E)-\mathtt{E}_E(s_E)\Vert_2^2\right] \\
    + \mathbb{E}_{\scaleto{s_A\sim \mathcal{D}_{\mathcal{M}_A^{\mathcal{T}_j}}}{10pt}}\left[\Vert\mathtt{E}_E\circ\phi(s_A)-\mathtt{E}_A(s_A)\Vert_2^2\right]
\end{multline}

\begin{figure*}[ht]
    \centering
    \includegraphics[scale=0.5]{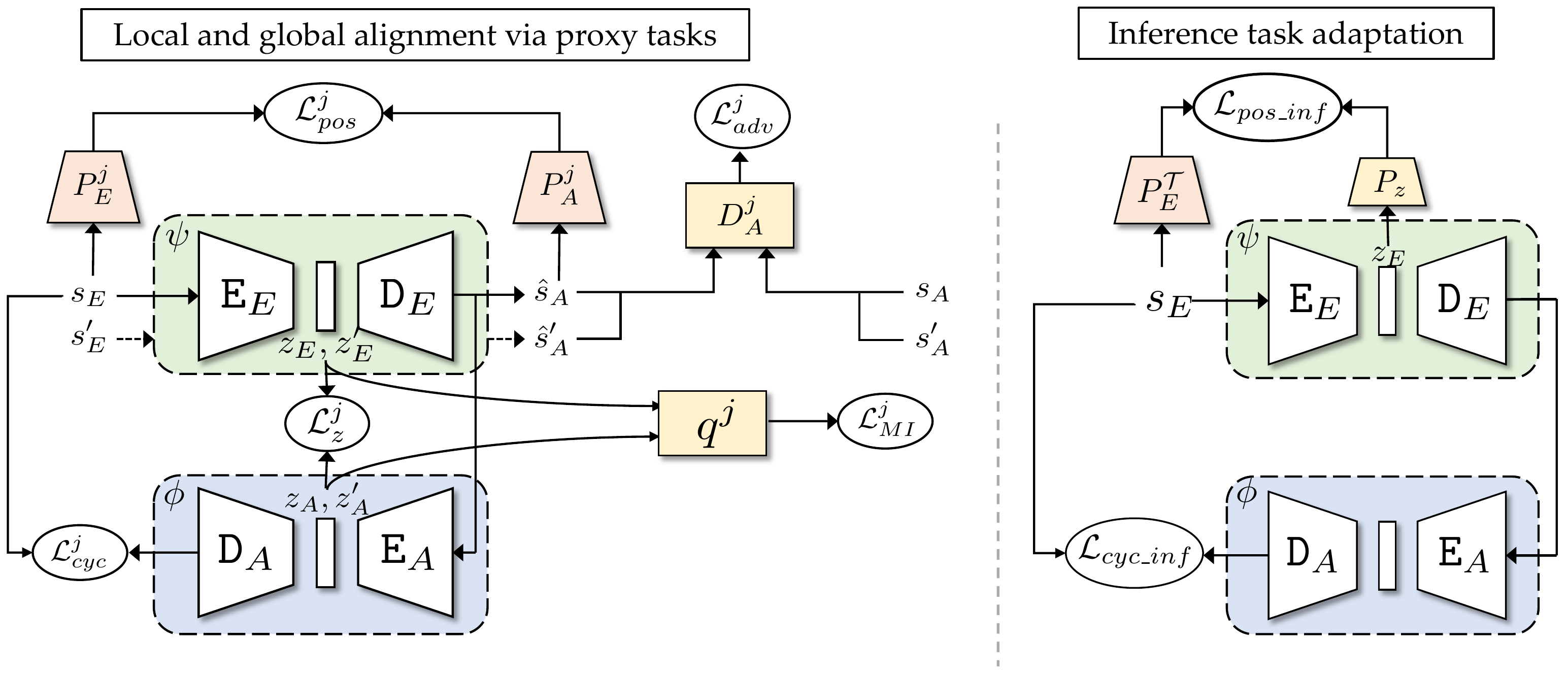}
    \vspace{-0.7em}
    \caption{\textbf{Framework overview.} An illustration of our MDP correspondence learning framework. We perform local alignment via state-transition distribution matching and cycle-consistency in the state space using $\mathcal{L}_{adv}^j$ and $\mathcal{L}_{cyc}^j$, as well as in a learnt latent space using $\mathcal{L}_{z}^j$ and $\mathcal{L}_{MI}^j$(only proxy task is $j$ shown here). The inverse cycle from agent to expert is omitted here for clarity. Global alignment is performed via consistency on the temporal position of states across the two domains, using the pre-trained position estimators $P_A^j,P_E^j$ in $\mathcal{L}_{pos}^j$. Further improvement is obtained via inference task adaptation using $\mathcal{L}_{pos\_inf}^j$ and $\mathcal{L}_{cyc\_inf}^j$ - this prevents overfitting to the proxy tasks and makes the learnt transformation more robust and well-conditioned to the target data.}
    \label{fig:framework}
    \vskip -0.1in
\end{figure*}

\subsection{Global alignment via temporal position preservation}
\label{sec:pos}
Solely learning with state cycle-consistency may result only in local alignment: an optimal state pair in the expert domain may get mapped to an optimal transition in the agent domain while violating task semantics (transitions from beginning of a task get mapped to terminal ones), and then back without breaking cycle-consistency. In order to constrain the mapping to maintain temporal semantics for a task, we enforce additional consistency on the temporal position of states across the two domains.

We encode the temporal position of a state by computing a normalized score of proximity to the terminal state in the trajectory. Each state is assigned a value of $1$ if they are terminating goal states and $0$ otherwise. These discrete values are then exponentially weighted by a discount factor $\gamma\in(0,1)$ to obtain a continuous estimate of the state temporal position. Using these temporal encodings, we pre-train temporal position estimators $P^j_E, P^j_A$ in a supervised fashion by optimizing a squared error loss as follows:
\begin{equation}\label{eq:val}
    \min_{P^j_E} \mathbb{E}_{\scaleto{\eta\sim \mathcal{D}_{\mathcal{M}_E^{\mathcal{T}_j}}}{10pt}} \sum_{t=1}^{H_\eta}\bigg( P^j_E(s^t_E) -  \gamma^{H_{\eta}-t}\bigg)^2
\end{equation}
$P^j_A$ is learnt in a similar fashion by optimizing Equation \ref{eq:val} with respect to the agent trajectories. These estimators are subsequently used to enforce temporal preservation as: 
\begin{multline}\label{eq:val_con}
    \min_{\psi,\phi} \mathcal{L}_{pos}^j = \mathbb{E}_{\scaleto{s_E\sim \mathcal{D}_{\mathcal{M}_E^{\mathcal{T}_j}}}{10pt}}\left[\Vert P^j_A\circ\psi(s_E)- P^j_E(s_E)\Vert_2^2\right] \\+ 
    \mathbb{E}_{\scaleto{s_A\sim \mathcal{D}_{\mathcal{M}_A^{\mathcal{T}_j}}}{10pt}}\left[\Vert P^j_E\circ\phi(s_A)-P^j_A(s_A)\Vert_2^2\right].
\end{multline}
Our temporal position estimators may be interpreted as state value functions: trajectories are from a greedy optimal policy with reward $1$ for terminal states, and $0$ otherwise.


\subsection{Inference task adaptation}
As discussed in Section \ref{sec:setting}, we are provided with the state-only trajectories $\mathcal{D}_{\mathcal{M}_E^\mathcal{T}}$ on solely the expert domain for the inference task $\mathcal{T}$. 
We propose to use these trajectories during the learning process as additional regularization, referred to as inference task adaptation. First, we enforce cycle consistency on the states:
\begin{equation}
    \min_{\psi,\phi} \mathcal{L}_{cyc\_inf} = \mathbb{E}_{\scaleto{s_E\sim \mathcal{D}_{\mathcal{M}_E^{\mathcal{T}}}}{9pt}}\left[\Vert\phi\circ\psi(s_E)-s_E\Vert_2^2\right].
\end{equation}
In addition, we also enforce temporal preservation in the latent space. Concretely, we first train a position estimator $P^{\mathcal{T}}_E$ by optimizing Equation \ref{eq:val} on the given trajectories as discussed in Section \ref{sec:pos}. We use the trained position estimator, along with a latent space position predictor $P_z$ to enforce temporal preservation by:
\begin{equation}
    \min_{\mathtt{E}_E,P_z} \mathcal{L}_{pos\_inf} = \mathbb{E}_{\scaleto{s_E\sim \mathcal{D}_{\mathcal{M}_E^{\mathcal{T}}}}{9pt}}\left[\Vert P_z\circ\mathtt{E}_E(s_E)-  P^\mathcal{T}_E(s_E)\Vert_2^2\right]. 
\end{equation}

\subsection{Optimization}
Given the alignment dataset $D$ containing trajectories from the $M$ proxy tasks, we first pre-train the temporal position estimators $\{(P_E^j,P_A^j)\}_{j=1}^M$ using Equation \ref{eq:val}. This is followed by adversarial training of the state maps $\psi,\phi$, where we use separate discriminators on the state space and latent space for each proxy task. 
The full objective is then:
\begin{align}\label{eq:opt}
    \min_{\psi,\phi} & \max_{\scaleto{\{D_E^{j}\},\{D_A^{j}\},\{q^j\}}{7pt}}\mathcal{L} = \sum_{j=1}^M\bigg[\lambda_1\bigg(\mathcal{L}_{adv}^{j}(D_A^{j}) + 
    \mathcal{L}_{adv}^{j}(D_E^{j})\bigg) \nonumber \\ &+\lambda_2\bigg(\mathcal{L}^j_{cyc}+\mathcal{L}^j_{z}\bigg) + \lambda_3\mathcal{L}^j_{pos} - \lambda_4\mathcal{L}^j_{MI}\bigg] \nonumber \\
    &+ \lambda_5\bigg(\mathcal{L}_{cyc\_inf}+\mathcal{L}_{pos\_inf}\bigg), 
\end{align}
where $\{\lambda_i\}_{i=1}^{5}$ denote hyper-parameters which control the contribution of each loss term. A pictorial description of the overall framework is shown in Figure \ref{fig:framework}.

\begin{table*}[ht]
\centering
\small
\caption{Cross-domain imitation performance of the policy learnt on transferred trajectories for inference tasks. All rewards are normalized by expert performance on corresponding task.}
\vskip 0.05in
\begin{tabular}{@{}lcccccc@{}}
\toprule
\textsc{Method} & \textsc{V-R2R} & \textsc{V-R2W} & \textsc{D-R2R} & \textsc{M-R2R} & \textsc{M-A2A} & \textsc{M-A2C} \\ \midrule
\textsc{IF} 
& $0.32\pm0.10$   
& $0.57\pm0.20$ 
& $0.48\pm0.30$      
& $0.61\pm0.23$      
& $0.09\pm0.08$      
& $0.00\pm0.00$   \\
\textsc{CCA}  
& $0.16\pm0.27$ 
& $0.86\pm0.30$
& $0.47\pm0.20$      
& $0.16\pm0.13$      
& $0.30\pm0.30$      
& $0.75\pm0.50$   \\
\textsc{CycleGAN}  
& $0.17\pm0.10$  
& $0.72\pm0.16$
& $0.13\pm0.02$      
& $0.12\pm0.06$      
& $0.22\pm0.20$      
& $0.80\pm0.28$   \\
\textsc{Ours} 
& $\mathbf{0.95\pm0.03}$  
& $\mathbf{0.93\pm0.01}$
& $\mathbf{0.99\pm0.02}$      
& $\mathbf{0.96\pm0.07}$      
& $\mathbf{0.78\pm0.08}$      
& $\mathbf{1.00\pm0.00}$  \\
\bottomrule
\end{tabular}
\vskip -0.05in
\label{tab:perf}
\end{table*}
\subsection{Imitation from observation}
We use the learned $\psi$ to map the states in the inference task expert demonstrations $\mathcal{D}_{\mathcal{M}_E^{\mathcal{T}}}$ to the agent domain. Given the set of transferred state-only demonstrations $\hat{\mathcal{D}}_{\mathcal{M}_A^{\mathcal{T}}}$, 
we can use any imitation from observation algorithm to learn the final policy. In this work, we follow the Behavioral Cloning from Observation (BCO) approach proposed in \cite{torabi2018behavioral}. BCO entails learning an inverse dynamics model $\mathcal{I}_A:\mathcal{S}_A\times\mathcal{S}_A\rightarrow\mathcal{A}_A$ to infer missing action information. First, we collect a dataset of state-action triplets $\mathcal{P}=\{(s^t_A,a^t_A,s^{t+1}_A)\}$ by random exploration. The inverse model is subsequently estimated by Maximum Likelihood Estimation (MLE) of the observed transitions  in $\mathcal{P}$. Assuming a Gaussian distribution over actions, this reduces to minimizing an $\ell2$ loss as follows,
\begin{align}
    \min_{\mathcal{I}_A} \sum_{\scaleto{(s^t_A,a^t_A,s^{t+1}_A)\in \mathcal{P}}{9pt}} \Vert a^t_A - \mathcal{I}_A(s^t_A,s^{t+1}_A)\Vert^2_2 
\end{align}
Next, the learnt inverse model is used to augment $\hat{\mathcal{D}}_{\mathcal{M}_A^{\mathcal{T}}}$ with agent specific actions. Finally, these action augmented trajectories are used to learn the final policy $\pi_A^\mathcal{T}$ via behavioral cloning. Note that our correspondence learning framework is agnostic to the imitation from observation algorithm used for learning the agent policy. The pseudo-code for training our framework is presented in Appendix \ref{app_pseudo}.


\section{Experiments}

In this section, we analyze the efficacy of our proposed method on the \texttt{xDIO} task. We adopt MuJoCo \cite{todorov2012mujoco} as the experimental test-bed and evaluate on several cross-domain tasks, along with a thorough ablation study of different modules in our overall framework. Implementation details are presented in Appendix \ref{App_implementation}. Code and videos are available at: \url{https://driptarc.github.io/xdio.html}.

\subsection{Tasks}
We use a total of 7 environments derived from the OpenAI Gym \cite{brockman2016openai}: 2-link Reacher, 3-link Reacher, Friction-modified 2-link Reacher, Third-person 2-link Reacher, 4-legged Ant, 6-legged Ant and HalfCheetah. We use the joint level state-action space for all environments. These are used to construct six cross-domain tasks:\\
\textbf{Dynamics-Reacher2Reacher (D-R2R)}: Agent domain is the 2-link Reacher and expert domain is the Friction-modified 2-link Reacher, created by doubling the friction co-efficient of the former. The proxy tasks are reaching for $M$ goals and the inference tasks are reaching for $4$ new goals, placed maximally far away from the proxy goals. Refer to Appendix \ref{App_env} for more details on goal placement.\\
\textbf{Viewpoint-Reacher2Reacher (V-R2R)}: Agent domain is the 2-link Reacher and expert domain is Third-person 2-link Reacher that has a ``third person'' view state space with a $180^\circ$ planar offset. Tasks are the same as D-R2R.\\
\textbf{Viewpoint-Reacher2Writer (V-R2W)}: Agent domain is the 2-link Reacher and expert domain is Third-person 2-link Reacher. The proxy tasks are same as D-R2R and the inference task is tracing a letter on a plane as fast as possible \cite{kim2020domain}. The inference task differs from the proxy tasks in two key aspects: the end effector must draw a straight line from the letter’s vertex to vertex and not slow down at the vertices.\\
\textbf{Morphology-Reacher2Reacher (M-R2R)}: Agent domain is the 2-link Reacher, while expert domain is the 3-link Reacher. Otherwise same as D-R2R. \\
\textbf{Morphology-Ant2Ant (M-A2A)}: Agent domain is the
4-legged Ant, while expert domain is the 6-legged Ant. Otherwise same as D-R2R. \\
\textbf{Morphology-Ant2Cheetah (M-A2C)}: Agent domain is the
HalfCheetah, while expert domain is the 4-legged Ant. Otherwise same as D-R2R.


\begin{figure}[h]
    \centering
    \vskip -0.1in
    \includegraphics[scale=0.45]{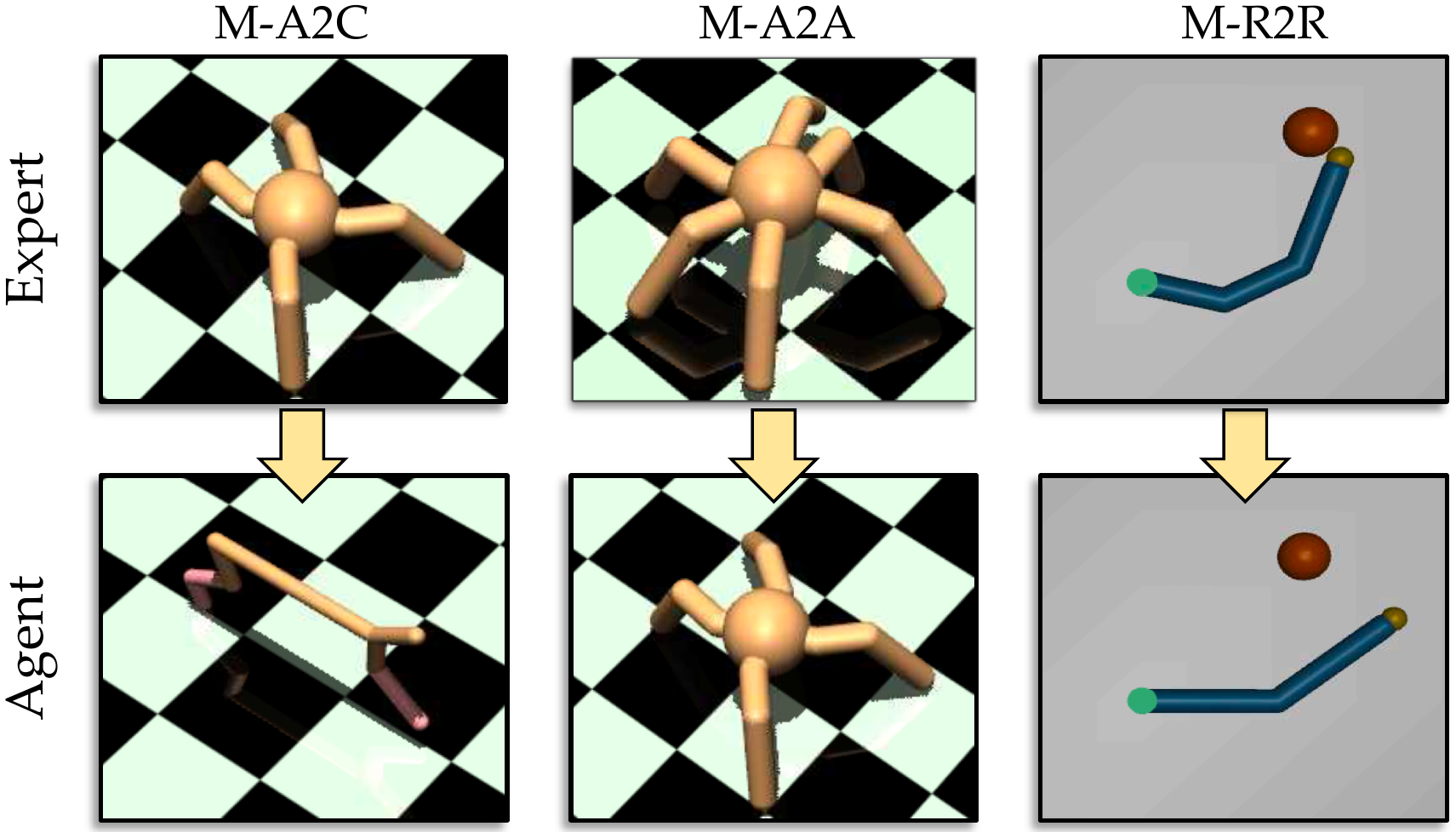}
    \caption{\textbf{Cross-domain tasks.} Different morphologically mismatched tasks used in our experiments.}
    \label{fig:tasks}
    \vskip -0.2in
\end{figure}
\begin{figure*}[ht]
    \centering
    \includegraphics[scale=0.65]{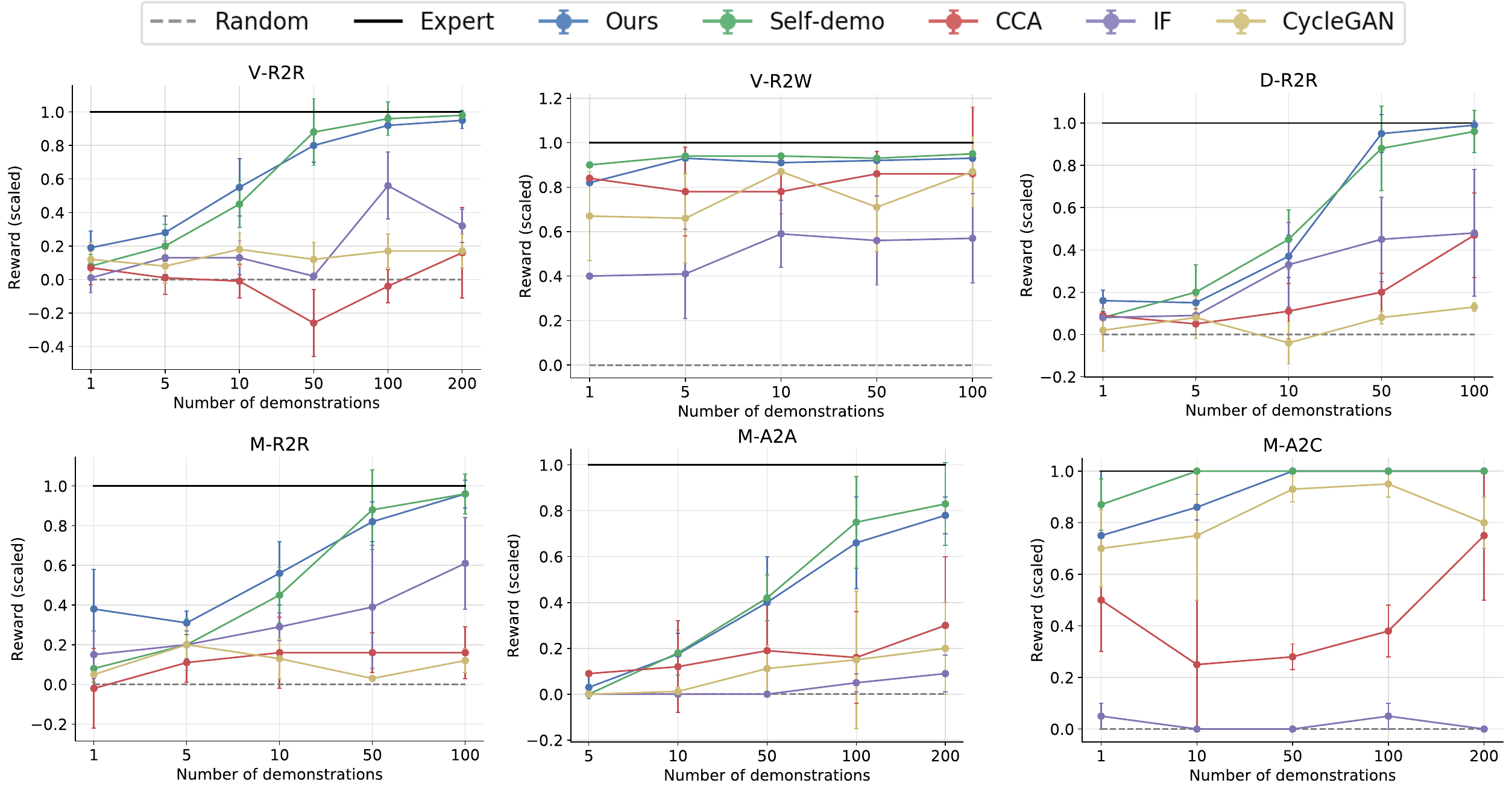}
    \caption{\textbf{Adaptation complexity.} Performance of learned policy as as the number of cross-domain demonstrations is varied. Our framework consistently performs better than baselines and achieves results close to Self-demo.}
    \label{fig:adapt}
    \vskip -0.1in
\end{figure*}

\begin{table*}[t]
\centering
\small
\caption{Ablation study on each module's contribution to final policy performance.}
\vskip 0.05in
\begin{tabular}{@{}lcccccc@{}}
\toprule
\textsc{Method} & \textsc{V-R2R} & \textsc{V-R2W} &\textsc{D-R2R} & \textsc{M-R2R} & \textsc{M-A2A} & \textsc{M-A2C} \\ \midrule
\textsc{Ours}                  
& $\mathbf{0.95\pm0.05}$
& $\mathbf{0.93\pm0.00}$
& $\mathbf{0.99\pm0.02}$      
& $\mathbf{0.96\pm0.07}$      
& $\mathbf{0.78\pm0.08}$      
& $\mathbf{1.00\pm0.00}$   \\
\ \textsc{- w/o Inference Adaptation}        
& $0.81\pm0.11$  
& $0.88\pm0.03$
& $0.74\pm0.22$      
& $0.78\pm0.11$      
& $0.46\pm0.12$      
& $0.78\pm0.23$   \\
\ \textsc{- w/o} $\mathcal{L}_{MI}$ 
& $0.60\pm0.30$   
& $0.92\pm0.03$
& $0.76\pm0.30$      
& $0.67\pm0.34$      
& $0.28\pm0.20$      
& $0.80\pm0.21$   \\
\ \textsc{- w/o Temporal Preservation} 
& $0.64\pm0.31$ 
& $0.84\pm0.00$
& $0.70\pm0.32$     
& $0.72\pm0.32$      
& $0.36\pm0.50$       
& $0.43\pm0.50$  \\
\bottomrule
\end{tabular}
\vskip -0.05in
\label{tab:ablation}
\end{table*}
\begin{figure*}[ht]
    \centering
    \includegraphics[scale=0.65]{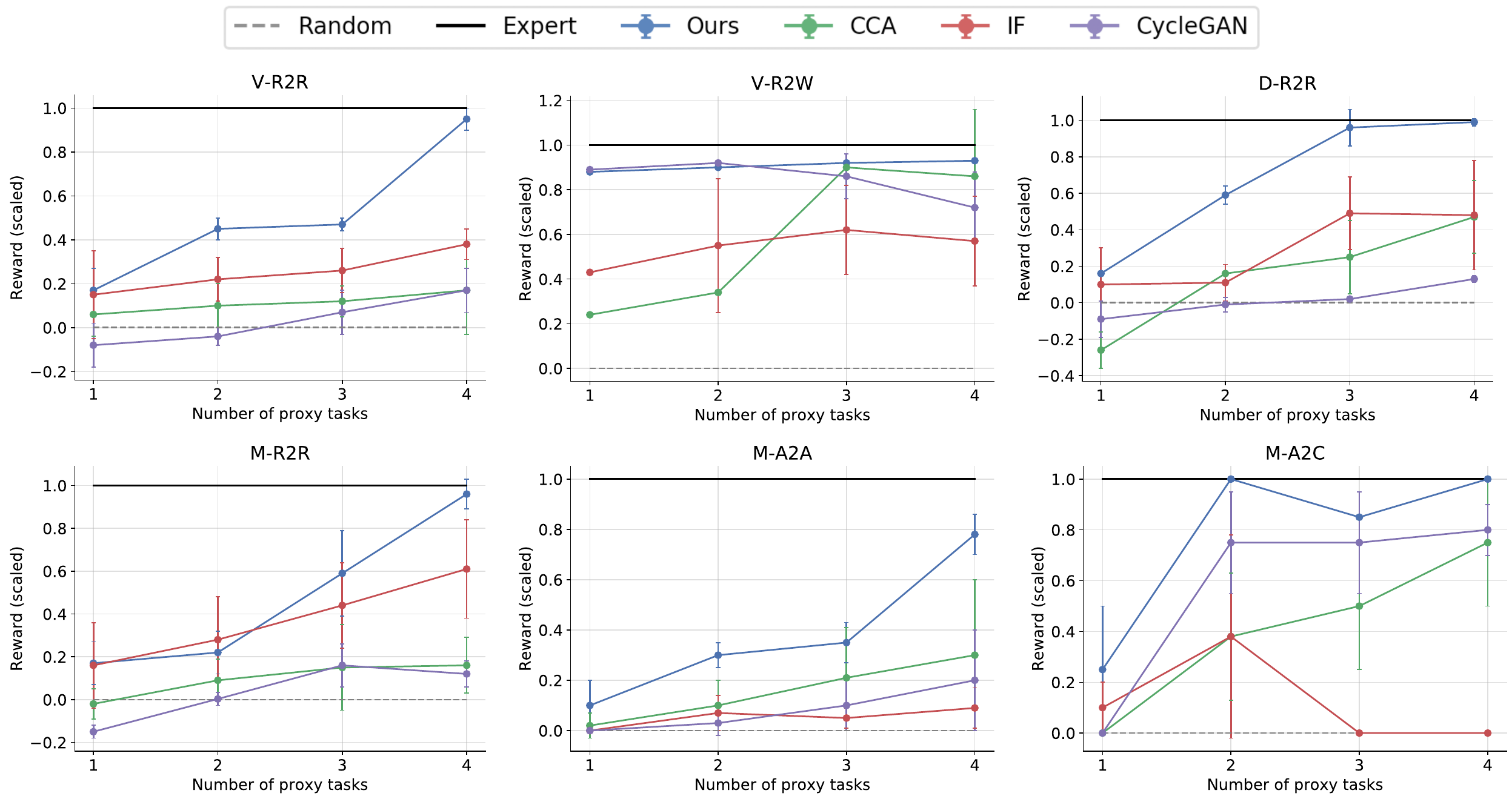}
    \caption{\textbf{Alignment Complexity.} Performance of learned policy as as the number of proxy tasks is varied. Notably, even with a reduced number of proxy tasks, our method outperforms the baselines in most cases.}
    \label{fig:align}
    \vskip -0.15in
\end{figure*}

\subsection{Baselines}
We compare our framework to other methods which are able to learn state correspondences from unpaired and unaligned demonstrations without access to expert actions - Canonical Correlation Analysis \cite{hotelling1992relations}, Invariant Features \cite{gupta2017invariant} and CycleGAN \cite{zhu2017unpaired}. Canonical Correlation Analysis (CCA) \cite{hotelling1992relations} finds invertible linear transformations to a space where domain data are maximally correlated when given unpaired, unaligned demonstrations. Invariant Features (IF) learns state maps via a domain agnostic space from paired and aligned demonstrations - we use Dynamic Time Warping \cite{muller2007dynamic} on the learned latent space to compute the pairings from the unpaired data. CycleGAN learns the state correspondence via adversarial learning with an additional cycle-consistency on state reconstruction. For all the baselines, we follow a similar procedure towards learning the final policy - the correspondence is learnt through the proxy tasks and then is used to transfer trajectories for policy training via BCO. Reported results are averaged across 10 runs. Experts on Reacher tasks are trained using PPO \cite{schulman2017proximal}, while those for Ant/Cheetah are trained using A3C \cite{mnih2016asynchronous}.

\begin{figure}[!ht]
    \centering
    \includegraphics[scale=0.5]{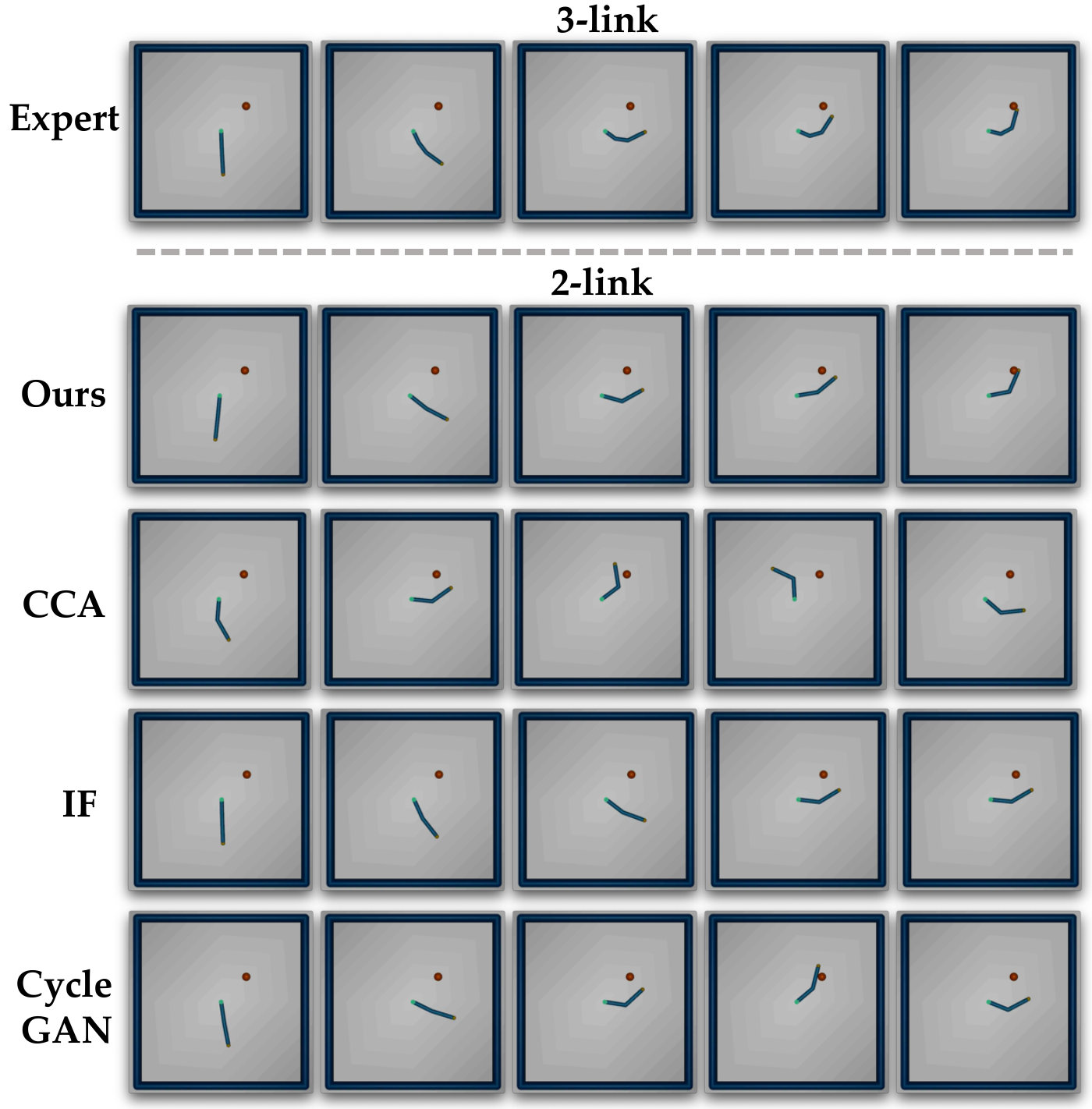}
    \caption{\textbf{Visualization of domain transformations.} State maps learned by our framework and the baselines on the M-R2R task. Our framework is able to map the end effector in a manner which preserves task semantics.}
    \label{fig:viz}
    \vskip -0.2in
\end{figure}

\subsection{Cross-domain imitation performance}
We compare imitation policies learnt by our framework against those learnt using baselines in Table~\ref{tab:perf}. As may be observed, the proposed method achieves near expert performance across all the cross-domain tasks encompassing viewpoint, dynamics and morphological mismatch. On the other hand, baselines consistently fail to generalize across the same tasks. There are two key reasons which can be hypothesized for this poor performance. Firstly, IF requires time-aligned trajectories, and the alignment when done by algorithms like DTW, rather than human intervention, may not be good enough given that our experiments involve diverse starting states, up to $1.5\times$ differences in demonstration lengths, and varying task execution rates. Secondly, baselines which learn from unpaired data (CCA and CycleGAN), also fail due to the lack of a mechanism to preserve MDP task characteristics, which is taken care of in our method via temporal order preservation and domain alignment. Figure \ref{fig:viz} illustrates the learnt state-maps for some of the cross-domain tasks. The proposed framework translates the expert states in a manner that preserves task semantics. 

\noindent \textbf{Varying the number of demonstrations.} 
Given an adequate set of proxy tasks, we experiment by varying the number of cross-domain demonstrations required for training the policy on the inference task. To serve as an upper-bound on performance, we imitate on agent domain demonstrations, drawn from an expert, on the inference task and denote this as the Self-demo baseline. As shown in Figure \ref{fig:adapt}, our framework produces transferred demonstrations of equal effectiveness to the self-demonstrations. This clearly demonstrates the effectiveness of our framework. 

\noindent \textbf{Varying the number of proxy tasks.}
The number of proxy tasks play a vital role in learning the correspondence across the domains. We perform experiments by varying the number of proxy tasks in the alignment set needed to learn the state-map for imitation, given sufficient cross-domain demonstrations for the inference tasks. The results are shown in Figure \ref{fig:align}. In general, more proxy tasks equate to better domain alignment as the solution space over possible state maps is constrained, and the learnt mapping generalizes better to the inference tasks.

\subsection{Ablation study}
We perform a set of ablation studies by removing each piece of the framework, demonstrating the importance of including each component. The results are shown in Table \ref{tab:ablation}. We begin by excluding inference task adaptation. This leads to a small drop in performance across all tasks, reinforcing the need for adapting on the inference task to incorporate the new state distribution introduced by the inference task. Notably, even without adaptation, the performance in almost all the tasks exceeds those of the baselines. Removing the mutual information objective leads to a similar drop in performance across all tasks. Excluding temporal position preservation also reduces performance -- demonstrating the significance of preserving task semantics via global alignment, which cycle-consistency alone fails to ensure.

\section{Conclusion}
In this paper, we present a novel framework to tackle the \texttt{xDIO} task by learning a state-map across domains using both local and global alignment. Local alignment is performed via transition distribution matching and cycle-consistency in both the state and latent space, while global alignment is enforced via the idea of temporal position preservation. While previous approaches rely on paired data and expert actions, we provide a general framework that can learn the mapping from unpaired, unaligned demonstrations without expert actions. We demonstrate the efficacy of our approach on multiple cross-domain tasks encompassing dynamics, viewpoint and morphological mismatch. Our future work will concentrate on extending our method for learning correspondence using random trajectories, thus mitigating the need for proxy tasks.

\section*{Acknowledgements}
This work was partially supported by Mitsubishi Electric Research Labs and National Institute of Food and Agriculture Award No: 2021-67022-33453 through the National Robotics Initiative.

\bibliography{ref.bib}
\bibliographystyle{icml2021}

\newpage

\appendix 
\clearpage

\section*{Appendix}
\section{Pseudo-code} \label{app_pseudo}
\begin{algorithm}[ht]
   \caption{Learn domain transformation $\psi$}
   \label{alg:correspondence}
\begin{algorithmic}
   \STATE {\bfseries Input:} Proxy task set $\scaleto{\left\{(\mathcal{D}_{\mathcal{M}_E^{\mathcal{T}_j}},\mathcal{D}_{\mathcal{M}_A^{\mathcal{T}_j}})\right\}_{j=1}^M}{20pt}$, inference task trajectories $\scaleto{\mathcal{D}_{\mathcal{M}_E^{\mathcal{T}}}}{11pt}$
   
   \WHILE{not done}
   
   \FOR[\hfill //Global and local alignment]{$j=1,\dots,M$} 
   
   \STATE Sample $\scaleto{(s_E,s_E')\sim\mathcal{D}_{\mathcal{M}_E^{\mathcal{T}_i}},(s_A,s_A')\sim\mathcal{D}_{\mathcal{M}_A^{\mathcal{T}_i}}}{12pt}$ and store in buffers $B_E^j,B_A^j$  
   
   \FOR{$i=1,\dots,N$}
   
   \STATE Sample mini-batch $i$ from $B_E^j,B_A^j$
   
   \STATE Update $D_E^j,D_A^j$ by maximizing $\mathcal{L}_{adv}^i(D_E^j)$ and $\mathcal{L}_{adv}^j(D_E^j)$ respectively
   
   \STATE Update $q^j$ by minimizing $\mathcal{L}_{MI}^j$
   
   \STATE Update $\psi,\phi$ by minimizing 
   $\lambda_1\left(\mathcal{L}_{adv}^{j}(D_A^{j}) + 
    \mathcal{L}_{adv}^{j}(D_E^{j})\right) + \lambda_2\left(\mathcal{L}^j_{cyc}+\mathcal{L}^j_{z}\right) + \lambda_3\mathcal{L}^j_{pos} - \lambda_4\mathcal{L}^j_{MI}$
   
   \ENDFOR
   
   \ENDFOR
   
   \STATE Sample $\scaleto{(s_E,s_E')\sim\mathcal{D}_{\mathcal{M}_E^{\mathcal{T}}}}{12pt}$ and store in buffers $B_E^{M+1}$ \COMMENT{ //Inference task adaptation}
   
   \FOR{$i=1,\dots,N$}
   
   \STATE Sample mini-batch $i$ from $B_E^{M+1}$
   
   \STATE Update $V_z$ by minimizing $\mathcal{L}_{pos\_inf}$
   
   \STATE Update $\psi,\phi$ by minimizing $\mathcal{L}_{cyc\_inf}+\mathcal{L}_{pos\_inf}$
   
   \ENDFOR
   
   \ENDWHILE
\end{algorithmic}
\end{algorithm}

\section{Implementation details} \label{App_implementation}

\textbf{Baselines.}
We use a total of $200$ expert trajectories for each proxy task, in both thee expert and self domains, to learn the state map. For IF, we use Dynamic Time Warping (DTW) \cite{muller2007dynamic} to obtain state correspondences. First, we randomly pair trajectories (due to lack of pairing) and simply pair the states that are visited in the same time step in the two proxy domains and use it to learn a common feature space.  This feature space serves as a metric space for DTW to re-estimate correspondences across domains. The new correspondences are then used as pairs for learning a better feature space, and so on. For CycleGAN, we follow an adversarial learning scheme similar to our framework, with consistency applied both on the state and latent spaces. To visualize and evaluate the state maps learned in prior work, we use the encoder and decoder for IF and the Moore-Penrose pseudo inverse of the embedding matrix for CCA.

\textbf{Architecture.} 
The state maps $\{\psi,\phi\}$ are neural networks, with hidden layers of size $[128,64]$ (both encoder and decoder), on the Reacher experiments and $[512,256]$ for the others. The state space discriminators $\{D_A^j,D_E^j\}_{j=1}^M$ and latent space discriminators $\{q^j\}_{j=1}^M$ comprises hidden layers of size $[128,128]$ for the Reacher experiments and $[512,256,128]$ for the rest. All discriminators use spectral normalization \cite{miyato2018spectral} and additionally, replace the negative log likelihood objective in $\mathcal{L}_{adv}$ by a least-squares loss \cite{mao2017least}. This loss has been shown to be more stable during training. Temporal position estimators $\{P_A^j,P_E^j\}_{j=1}^M$ consist of hidden layers of size $[200,128]$. Latent space position estimator $V_z$, for the inference task adaptation, contain hidden layers of size $[64,64]$. The fitted policy $\pi_A^{\mathcal{T}}$ and the inverse dynamics model $\mathcal{I}_A$ have hidden layers of size $[64,64]$ and $[100,100]$ respectively. For CycleGAN, we use the same architecture as the state map in our framework. For IF, we use hidden layers with $[128,64]$ units and leaky ReLU non-linearities to parameterize the encoders and decoders. We use Adam optimizer with default decay rates and learning rate $1e\text{-}4$ for training. With regards to the hyperparameters in Eqn. 10, we set them as For our experiments, $\lambda_1=2,\lambda_2=\lambda_3=\lambda_4=\lambda_5=1$. Finally, for CCA, the embedding dimension is the minimum state dimension between the expert and self domains. We train all our models on a single Titan Xp GPU using PyTorch.
\begin{figure}[b]
    \centering
    \includegraphics[scale=0.45]{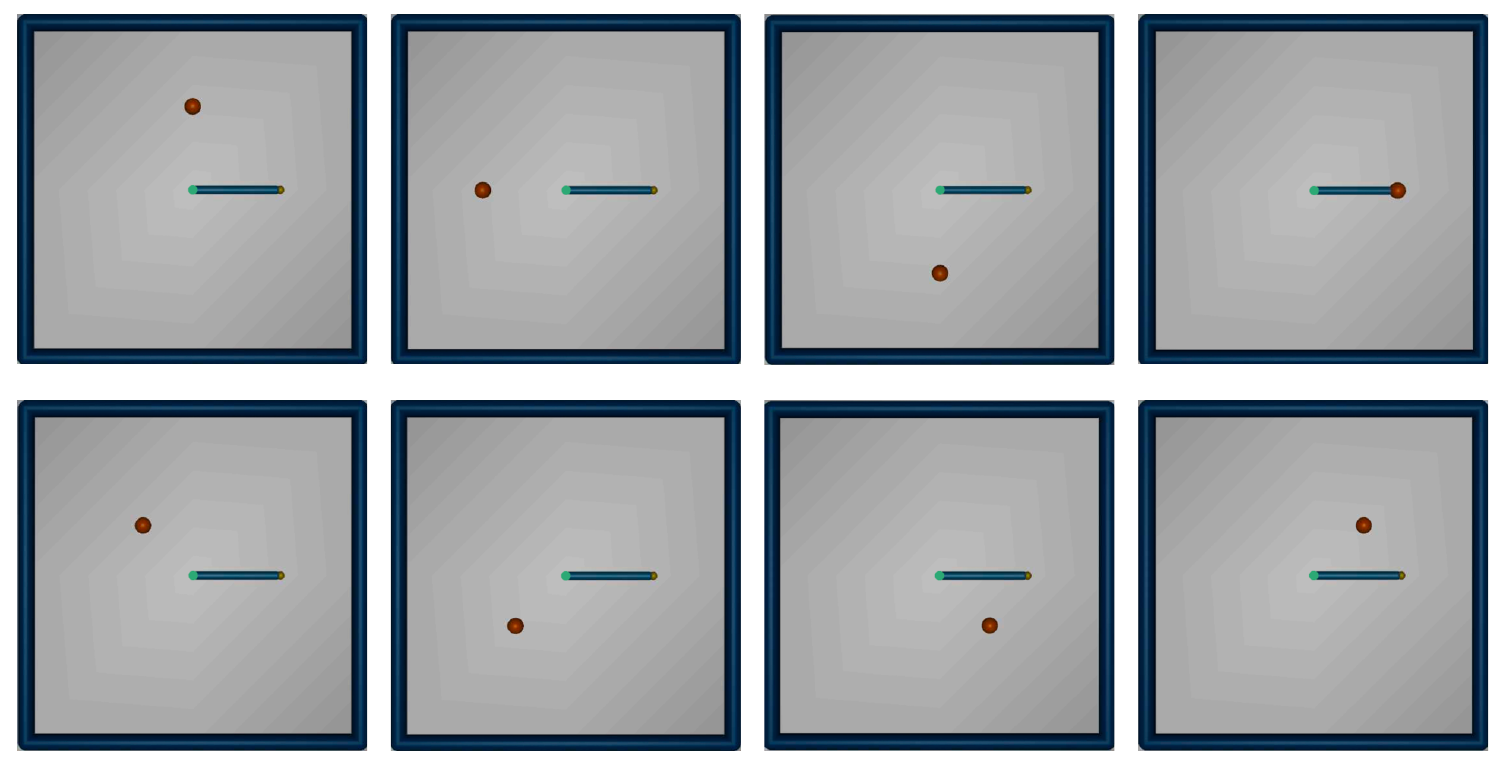}
    \caption{\textbf{Reacher task visualization.} The goal locations used in our reacher experiments. Top four goals constitute the set of proxy tasks, the bottom four serve as inference goals.}
    \label{fig:goals_reacher}
    \vskip -0.1in
\end{figure}

\begin{figure}[t]
    \centering
    \includegraphics[scale=0.45]{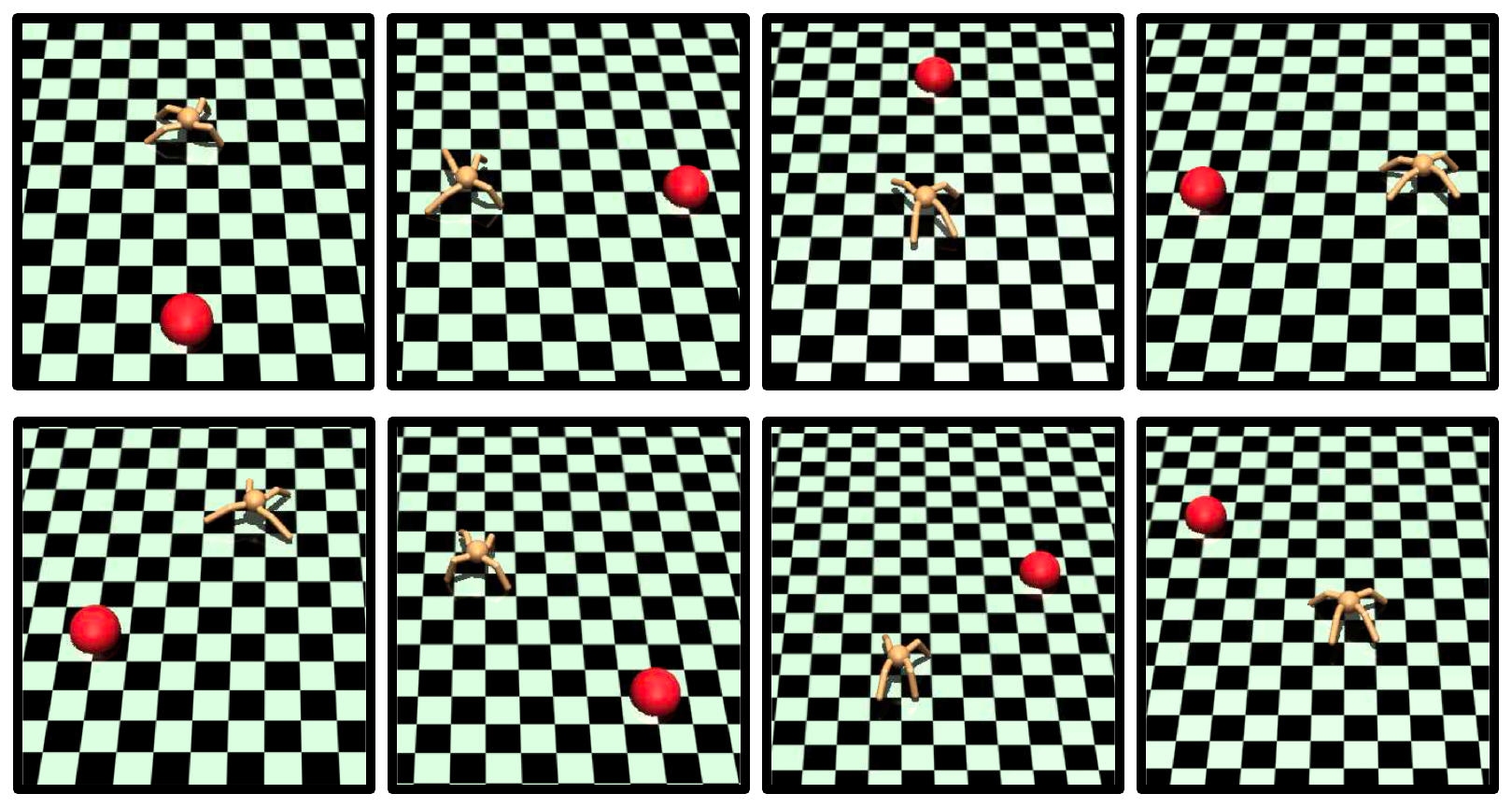}
    \caption{\textbf{Ant task visualization.} The tasks used in our ant and cheetah experiments. Top four constitute the set of proxy tasks, the bottom four serve as inference tasks.}
    \label{fig:goals_ant}
    \vskip -0.1in
\end{figure}
\section{Environment details} \label{App_env}
The various reacher environments used in the tasks are extended from the ``Reacher-v2'' OpenAI Gym \cite{brockman2016openai} environment. A $k$ link reacher has a state vector of the form $(\omega_1,\dots,\omega_k,\dot\omega_1,\dots,\dot\omega_k,x_g,y_g)$, where $\omega_i$ and $\dot \omega_i$ are the joint angle and angular velocity of the $i$th joint, and $(x_g,y_g)$ is the position of the goal. The action vector has the form $(\tau_1,\dots,\tau_k)$, where $\tau_i$ is the torque applied to the $i$th joint. The state map acts only on the non-goal dimensions. Following \cite{kim2020domain}, proxy goals are placed near the wall of the arena and the target tasks are reaching for 4 new goals near the corner of the arena. The new goals are placed as far as possible from the proxy goals within the bounds of the arena. Figure \ref{fig:goals_reacher} depicts the location of the goals.

\begin{wrapfigure}{r}{0.15\textwidth}
  \begin{center}
    \includegraphics[scale=0.3]{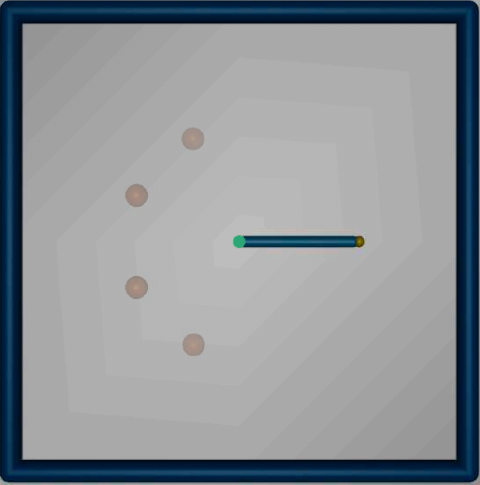}
  \end{center}
  \vspace{-0.8em}
  \caption{\textbf{V-R2W inference task.} The sequence of goals need to be reached as quick as possible.}
  \label{fig:goal_writer}
\end{wrapfigure}

For the V-R2W task, the proxy tasks are the same as the ones discussed previously, while the target task is tracing the letter C (shown in Figure \ref{fig:goal_writer}) as fast as possible. The goal location in the writing task represents the next vertex of the letter to trace. Once the first vertex is reached, the goal coordinates are updated to be the next vertex coordinates. The reward function is defined as follows:
\[
    R_{write}(s)= 
\begin{cases}
    100,& \text{if state $s$ corresponds to reaching a vertex} \\
    -1,              & \text{otherwise}
\end{cases}
\]
Thus the agent must perform a sequential reaching task and accomplish it as fast as possible. The key difference with a normal reaching task is that the reacher must not slow down at each vertex and plan its path accordingly in order to minimize drastic direction changes.

The two ant environments and the cheetah environment are derived derived from the ``Ant-v2'' and ``HalfCheetah-v2'' environments respectively. A $k$-legged Ant has a state vector of the form  $(c_x,c_y,c_z,q_0,\dots,q_3,\omega_1,\dots,\omega_{2k},\dot c_x,\dot c_y,\dot c_z,\allowbreak\dot q_1,\dot q_2,\dot q_3,\dot\omega_1,\dots,\dot\omega_{2k},x_g,y_g)$, where $(c_x,c_y,c_z)$ denotes the torso 3D co-ordinates, $(q_0,\dots,q_3)$ denotes the torso orientation quarternion, $(\dot c_x, \dot c_y, \dot c_z)$ denotes the torso 3D velocity and $(\dot q_1,\dot q_2,\dot q_3)$ denotes the torso angular velocity. The rest are the same as the reacher, with 2 hinge joints per leg. The action vector has the form $(\tau_1,\dots,\tau_{2k})$, where $\tau_i$ is the torque applied to the $i$th joint. For the cheetah, the state vector is of the form $(r_x,r_y,r_z,\omega_1,\dots,\omega_6,\dot r_x,\dot r_y,\dot r_z,\dot \omega_1,\dots,\dot \omega_6,x_g,y_g)$ where $(r_x,r_y,r_z)$ denotes the root 3D co-ordinates and $(\dot r_x,\dot r_y,\dot r_z)$ are the corresponding velocities; rest are the same as the reacher for the 6 hinge joints (3 for each leg). The action vector has the form $(\tau_1,\dots,\tau_6)$, where $\tau_i$ is the torque applied to the $i$th joint. For all these environments, the task is to reach the center of a circle of radius 5m with the agent being initialized on a $2^\circ$ arc of the circle. Different initializations define the different tasks as shown in Fig. \ref{fig:goals_ant}.

\end{document}